# A Low-Rank Method for Vision Language Model Hallucination Mitigation in Autonomous Driving

Keke Long[1], Jiacheng Guo[2], Tianyun Zhang[2], Hongkai Yu[2*], Xiaopeng Li[1*]

[1] University of Wisconsin – Madison, USA

[2] Cleveland State University, USA

Corresponding author: Hongkai Yu (h.yu19@csuohio.edu), Xiaopeng Li (xli2485@wisc.edu)

**Abstract:** Vision Language Models (VLMs) are increasingly used in autonomous driving to help understand traffic scenes, but they sometimes produce hallucinations, which are false details not grounded in the visual input. Detecting and mitigating hallucinations is challenging when ground-truth references are unavailable and model internals are inaccessible. This paper proposes a novel self-contained low-rank approach to automatically rank multiple candidate captions generated by multiple VLMs based on their hallucination levels, using only the captions themselves without requiring external references or model access. By constructing a sentence-embedding matrix and decomposing it into a low-rank consensus component and a sparse residual, we use the residual magnitude to rank captions—selecting the one with the smallest residual as the most hallucination-free. Experiments on the NuScenes dataset demonstrate that our approach achieves 87% selection accuracy in identifying hallucination-free captions, representing a 19% improvement over the unfiltered baseline and a 6-10% improvement over multi-agent debate method. The sorting produced by sparse error magnitudes shows strong correlation with human judgments of hallucinations, validating our scoring mechanism. Additionally, our method, which can be easily parallelized, reduces inference time by 51-67% compared to debate approaches, making it practical for real-time autonomous driving applications.

# 1 INTRODUCTION

Vision-language models (VLMs) are increasingly deployed in the transportation domain to interpret complex traffic scenes to support decision-making in autonomous driving (Guo et al., 2024; Qian et al., 2024). However, these models frequently produce hallucinations, which are plausible-sounding but factually incorrect details that are not grounded in the visual input (Farquhar et al., 2024). Figure 1 illustrates two typical types of hallucinations in driving scenarios. Such hallucinations pose critical safety risks in traffic scenarios: a VLM incorrectly describing a "stopped vehicle" as "moving" or hallucinating non-existent pedestrians could trigger inappropriate vehicle responses and potentially catastrophic consequences. Therefore, mitigating VLM hallucinations is essential for their safe deployment in autonomous driving.

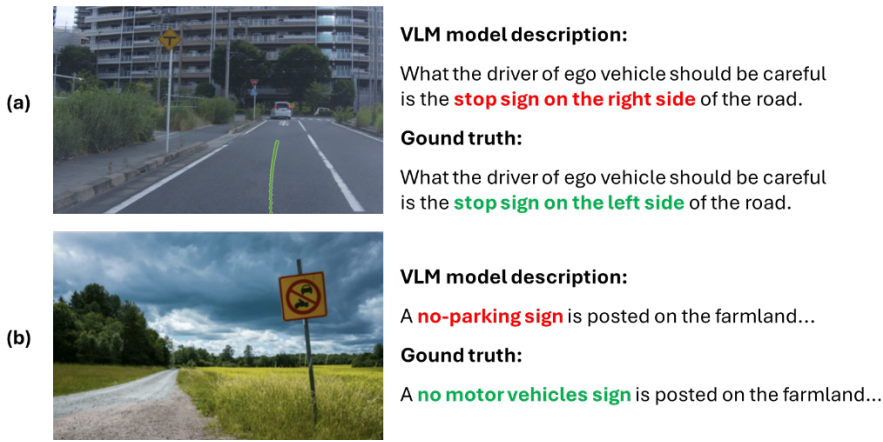

Figure 1 Examples of VLM hallucinations in driving scenarios (a):(Arai et al., 2025) (b):(Qian et al., 2024).

Hallucinations in VLMs can be addressed at three stages: data, model, and inference (Bai et al., 2025; Huang et al., 2025; Zhang et al., 2025). However, for application developers in the autonomous driving domain, focusing on inference-stage mitigation is both necessary and pragmatic. Most state-of-the-art VLMs are either proprietary black-box models accessible only through APIs (e.g., GPT-4V) or require prohibitive computational resources for training or adaptation (Karmanov et al., n.d.). Furthermore, the dynamic nature of traffic environments—with varying weather conditions, regional differences, and continuously evolving scenarios—makes it impractical to maintain specialized models for every context (Li et al., 2024). Therefore, hallucination mitigation methods during inference that can work with existing VLM models offer the most viable path toward reducing hallucinations in production systems. Yet, despite this clear need, inference-stage hallucination mitigation solutions remain largely unexplored in the autonomous driving domain.

To address this gap, we examine existing inference-based mitigation methods from the broader VLM literature, which can be categorized into three classes based on their information requirements, summarized in TABLE 1. Methods requiring model internals leverage information such as token probabilities, attention weights, or uncertainty estimates to identify potentially hallucinated content (Azaria and Mitchell, 2023; K. Li et al., 2023). While effective, these approaches are incompatible with commercial APIs that only provide final outputs, limiting their practical applicability. Methods Requiring External Knowledge validate model outputs against external sources such as knowledge bases (X. Li et al., 2023), retrieval systems (Gao et al., 2023), or structured databases. In the autonomous driving context, this might include checking against high-definition maps, traffic rule databases, or historical sensor data. However, maintaining comprehensive and up-to-date knowledge sources is resource-intensive, and coverage gaps are inevitable—especially for unusual scenarios or newly developed areas.

Self-contained methods operate solely on the model outputs without requiring internal model information or external knowledge. Representative approaches include vote (Z. Wang et al., 2024) and debate (Wang et al., 2023). The voting method generates multiple response candidates for a given question and selects the most frequently occurring answer as the final output. This mechanism has proven to be a simple yet effective strategy across diverse training paradigms. However, majority voting relies on exact matching, limiting its applicability to closed-end tasks such as arithmetic problems and multiple-choice questions. For open-ended vision descriptions—which constitute the majority of real-world autonomous driving scenarios—exact matching becomes infeasible as different captions may describe the same scene using varied but equally valid expressions (Jiang et al., 2025).

Multi-agent debate method approaches attempt to address this limitation by having multiple models engage in iterative discussion to reach consensus through argumentation and refinement (Y. Du et al., 2024). While this method can handle open-ended problems, it suffers from prohibitive computational overhead: each debate round requires multiple VLM inference calls, and given that a single VLM inference can take several seconds, the cumulative latency of multiple debate iterations makes it unsuitable for real-time applications where decisions must be made within milliseconds.

TABLE 1 Self-contained Methods VLM/LLM Hallucinations mitigation during Inference

| Category | Paper | Method | Task |
| --- | --- | --- | --- |
| Methods Requiring VLM/LLM Model Internals | (Azaria and Mitchell, 2023) | Statement accuracy prediction | Classification, Judgment (Y/N) |
| | (K. Li et al., 2023) | Inference-time intervention | Judgment (Y/N), QA |
| | (Chuang et al., 2024) | Contrast LLM's inter-layer logits during decoding. | Math questions and QA with fixed answers, QA without fixed answers |
| Methods Requiring External Knowledge | (X. Li et al., 2023) | Chain-of-Knowledge: use external text and structured data to guide LLMs. | Reasoning, Question and Answering (QA) |
| | (Gao et al., 2023) | Use web search to find supporting evidence and revise LLM outputs. | Factoid statements, reasoning chains, knowledge-intensive dialogs |
| | (Peng et al., 2023) | Use web search to find external evidence and revise LLM outputs. | Factoid statements, reasoning chains, knowledge-intensive dialogs |
| Self-contained Methods | (Wang et al., 2023) | Single-agent debate | QA tasks with unique answers |
| | (Z. Wang et al., 2024) | Single-agent vote | Multiple choice questions, true/false questions, and essay questions with unique answers |
| | (Y. Du et al., 2024) | Multi-agent debate | Arithmetic, grade-school math, chess moves, multiple-choice factual QA |
| | (Cohen et al., 2023) | Multi-agent debate | Fact completion, QA tasks with unique answers |
| | (Liang et al., 2024) | Multi-agent debate | Conversation and puzzle tasks with fixed answers, Trivia creative writing |
| | (Lin et al., 2024) | Multi-agent debate | Arithmetic, Translation |
| | (Yang et al., 2025) | Multi-agent debate and vote | Multiple-choice questions with fixed answers, open-ended questions requiring generating numerical values or reasoning |

Therefore, there is a critical need for methods that can effectively mitigate hallucinations in open-ended vision-language tasks without requiring lengthy iterative processes—combining the broad applicability needed for diverse traffic scenarios with the efficiency demanded by real-time autonomous systems.

This paper introduces a novel self-contained approach that addresses the limitations of existing methods through low-rank matrix decomposition of caption embeddings. Our key insight is that accurate descriptions of the same scene tend to share consistent semantic patterns that manifest as low-rank structures in the embedding space (Chen et al., 2024), while hallucinated content appears as sparse, high-dimensional deviations from this consensus (X. Du et al., 2024). Given multiple candidate captions generated by single or multiple VLMs, we construct a matrix of sentence embeddings and apply Singular Value Decomposition (SVD) to separate it into a low-rank consensus matrix capturing shared semantic information and a sparse residual matrix representing inconsistencies and potential hallucinations. By measuring the magnitude of residual components for each caption, we identify and select the caption with the smallest residual as the most reliable description with minimal hallucination.

The proposed approach is validated on the NuScenes dataset (Caesar et al., 2020), a large-scale autonomous driving dataset with complex traffic scenarios. Recent advances in autonomous driving have increasingly integrated VLMs as core perception components for scene understanding, hazard identification, and decision-making support (Long et al., 2024). Therefore, detecting and mitigating VLM hallucinations in this domain serves both as a rigorous stress test for our method and as a demonstration of practical value in a high-stakes real-world application.

The proposed method offers three contributions for autonomous driving applications:

1. We introduce the first low-rank framework to mitigate hallucinations in autonomous driving scene descriptions through multi-VLM collaborative consensus.
2. The proposed approach operates solely on VLM-generated captions, making it compatible with any black box VLM and deployable without ground truth.
3. The parallelizable architecture achieves sub-second hallucination mitigation, meeting real-time requirements critical for autonomous driving systems.

The remainder of this paper is organized as follows: Section 2 formally defines the problem, and Section 3 presents our low-rank decomposition framework. Section 4 describes the experimental setup and evaluation methodology. Section 5 presents comprehensive results on a real scenario dataset. Finally, Section 6 concludes with discussions of limitations and future research directions.

## 2 PROBLEM STATEMENT

We consider an image or video frame and aim to produce $n$ distinct descriptions or captions. Let $C = \{c_1, c_2, \ldots, c_n\}$ be the set of all candidate captions, where each $c_i$ is generated by one or more vision-language models (VLMs) via:

$$c_i = f^{\text{VLM}}(I, P_i), i \in \{1, 2, \cdots, n\} \quad (1)$$

where $P_i$ may vary slightly for each query (e.g., different prompts, sampling parameters, or different VLM models). In the NuScenes setting, for instance, our prompt $P_i$ focuses on vehicles and key traffic participants rather than background details.

Once we have the set $C$, our key question is how to identify captions $c_i \in C$ that contain the least hallucinated content. Our approach ranks these captions by analyzing their consistency within the embedding space, without requiring ground-truth references or human annotations.

## 3 METHODOLOGY

The proposed approach, identifying the least hallucinated caption, is illustrated in Figure 2. VLM agents generate candidate captions for the same image, which are then converted into embedding vectors. We construct a matrix from these embedding vectors and apply low-rank approximation to a consensus matrix and a hallucination matrix. Captions that deviate less from the low-rank consensus receive lower hallucination scores, enabling us to rank and select the most reliable caption. The following sections detail each component of this framework.

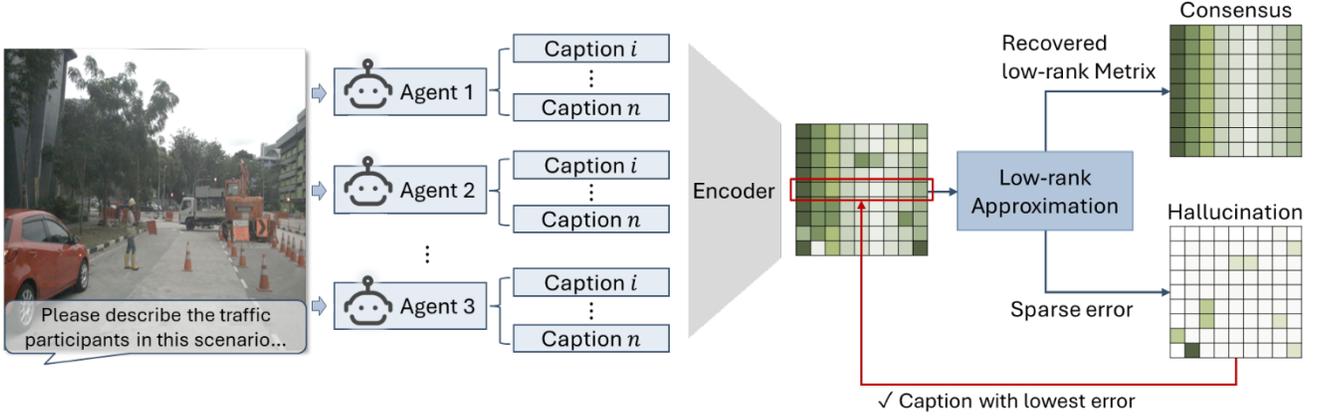

Figure 2 Basic structure of low-rank approximation method.

### 3.1 Sentence Embeddings

Given the set of captions $C$, we convert each caption $c_i$ into a $d$-dimensional embedding vector using a pre-trained sentence transformer:

$$e_i = \text{Enc}(c_i) \in \mathbb{R}^d, i = 1, \ldots, n \quad (2)$$

We then stack these embeddings row-wise into an $n \times d$ matrix:

$$\mathbf{M} = \begin{pmatrix} e_1^\mathrm{T} \\ e_2^\mathrm{T} \\ \vdots \\ e_n^\mathrm{T} \end{pmatrix} \in \mathbb{R}^{n \times d} \quad (3)$$

This matrix $\mathbf{M}$ serves as the input to our low-rank analysis framework.

### 3.2 Low-Rank Approximation

The core hypothesis is that captions describing consistent, accurate content will cluster in a low-dimensional subspace of the embedding space, while hallucinated captions will deviate significantly from this consensus representation. We then seek to recover a low-rank matrix $\mathbf{R}$ and approximate the sparse noises $\mathbf{E}$ from the matrix $\mathbf{M}$:

$$(\mathbf{R}^*, \mathbf{E}^*) = \arg\min_{\mathbf{R}, \mathbf{E}} (rank(\mathbf{R}) + \beta \|\mathbf{E}\|) \quad (4)$$

$$\text{s.t. } \mathbf{M} = \mathbf{R} + \mathbf{E}$$

where $\beta > 0$ is a balance factor between the rank of $\mathbf{R}$ and the norm of the noise $\mathbf{E}$. To capture this intuition, we employ Truncated SVD to find the best low-rank approximation of the embedding matrix. Given the embedding matrix $\mathbf{M} \in \mathbb{R}^{n \times d}$, we compute its truncated SVD with rank $r$:

$$\mathbf{M} \approx U_r \Sigma_r V_r^\mathrm{T} = \mathbf{R} \quad (5)$$

where $U_r \in \mathbb{R}^{n \times r}$ contains the first $r$ left singular vectors, $\Sigma_r \in \mathbb{R}^{r \times r}$ is a diagonal matrix with the top $r$ singular values, $V_r \in \mathbb{R}^{d \times r}$ contains the first $r$ right singular vectors, and $\mathbf{R} \in \mathbb{R}^{n \times d}$ is the rank-$r$ reconstruction of $\mathbf{M}$.

#### 3.2.1 Adaptive Rank Selection

An adaptive selection strategy based on cumulative variance explanation ratio is adopted to determine the rank value $r$. The core idea is to choose the minimal rank that captures sufficient total variance in the data. Specifically, we first compute the full SVD of $\mathbf{M}$ to obtain all singular values $\sigma_1 \geq \sigma_2 \geq \cdots \geq \sigma_{\min(n,d)}$. We then calculate the cumulative variance explanation ratio for each potential rank $k$:

$$\rho_k = \frac{\sum_{i=1}^{k} \sigma_1^2}{\sum_{q=1}^{\min(n,d)} \sigma_q^2} \quad (6)$$

The optimal rank $r$ is selected as the first value of $k$ such that $\rho_k \geq 0.95$, ensuring that the low-rank approximation preserves 95% of the total variance. This adaptive approach ensures that **R** captures the most dominant patterns shared across captions while remaining robust to datasets of varying complexity.

*3.2.2  Choice of Low-Rank Approximation Method: SVD vs RPCA*

While Robust Principal Component Analysis (RPCA) has shown effectiveness in computer vision applications for separating low-rank structures from sparse outliers (Yu et al., 2018), the choice between SVD and RPCA fundamentally depends on the error characteristics and the corresponding norm selection. The key distinction lies in their optimization objectives: SVD minimizes the Frobenius norm (L2) of the error, while RPCA minimizes a combination of nuclear norm of **R** and L1 norm of **E.** The different norms lead to fundamentally different behaviors. The L1 norm in RPCA promotes sparsity in the error matrix E—it assumes that corruptions affect only a small subset of entries, making it ideal for handling sparse outliers or pixel-level corruptions in images. In contrast, the Frobenius norm (L2) used by SVD distributes errors more evenly across all entries, making it suitable for dense, Gaussian-like noise patterns.

In our caption embedding analysis, the "noise" arises from natural linguistic variations rather than sparse corruptions. When captions are transformed into dense embedding vectors through pre-trained language models, semantic variations manifest as distributed differences across multiple dimensions rather than isolated outliers. Some paraphrases or synonym substitutions are not hallucinations but will create systematic shifts in the embedding space that affect many dimensions simultaneously. The L2 norm's tendency to minimize overall error magnitude while allowing it to be distributed across dimensions aligns perfectly with this characteristic. RPCA's L1 norm, which encourages exact zeros in the error matrix, would incorrectly treat these natural variations as sparse anomalies to be eliminated rather than distributed noise to be minimized.

From a practical standpoint, SVD offers significant advantages in computational efficiency and numerical stability. SVD decomposition has a well-defined computational complexity of $O(\min(mn^2, m^2n))$ and provides deterministic, unique solutions. In contrast, RPCA requires higher computational overhead, parameter tuning requirements.

Therefore, we employ truncated SVD with Frobenius norm minimization, which directly captures the principal semantic directions in the embedding space, aligning well with the distributed nature of linguistic variations, and it is computationally efficient to measure the residual.

*3.2.3  Residual Computation and Hallucination Scoring*

After obtaining the low-rank reconstruction $\mathbf{R}^*$, we compute the residual matrix:

$$\mathbf{E}^* = \mathbf{M} - \mathbf{R}^* \in \mathbb{R}^{n \times d} \quad (7)$$

Each row $E_{i,:}$ represents the deviation of caption $i$ from the low-rank consensus. We define the hallucination score for the caption $i$ as:

$$h_i = \|E_{i,:}^*\|_2 = \sqrt{\sum_{k=1}^{d} E_{i,k}^*} \quad (8)$$

A higher $h_i$ indicates that caption $c_i$ contains features that are not well-represented by the dominant subspace, suggesting potential hallucinations or inconsistencies. Conversely, a lower $h_i$ suggests that the caption is more consistent with the overall consensus.

Given the hallucination score $h_i$, the captions could be sorted in ascending order, and the one with the minimum score is selected as the most reliable output.

## 4 EXPERIMENTS

### 4.1 Experimental Setup

#### 4.1.1 Datasets

Our experiments utilize the NuScenes v1.0-trainval dataset, an autonomous driving scene dataset that provides comprehensive multi-modal sensor data. We select 300 keyframes, including various scenarios (intersections, ork zones, parking lots) and traffic participants (vehicles, pedestrians, cyclists), as shown in Figure 3. While the original NuScenes images have a resolution of 1600×900 pixels, we compress them to approximately 400×300 pixels to simulate more challenging perception tasks since lower image quality is more likely to induce hallucinations in VLMs.

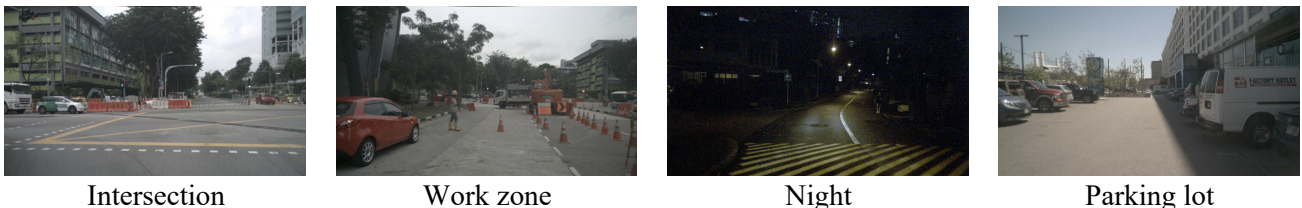

| Intersection | Work zone | Night | Parking lot |

Figure 3 Example of scene keyframes from NuScenes dataset

#### 4.1.2 Caption Generation Process

##### 4.1.2.1 Selected Vision-Language Models

We evaluate our approach using three representative vision-language models, shown in TABLE 2. Note that the goal of this work is not to benchmark the hallucination performance of existing vision-language models, but rather to propose and validate a low-rank framework for hallucination mitigation. To better highlight hallucination phenomena, we adopted small models and adjusted generation parameters (e.g., temperature) to increase output diversity. Therefore, the results should not be interpreted as the optimal performance of the tested models.

TABLE 2 Vision-language models list

| Model | Parameters | Publisher | Architecture |
|---|---|---|---|
| deepseek-vl2-tiny (Lu et al., 2024) | 1.3B | DeepSeek | Vision Transformer + LLM |
| llava-1.5-7b-hf (Liu et al., 2024) | 7B | Microsoft/UW | Vision-Language Alignment |
| Qwen2-VL-7B-Instruct (P. Wang et al., 2024) | 7B | Alibaba | Multi-modal Transformer |

##### 4.1.2.2 Prompt Design and Rationale

For NuScenes scenarios, we employ a carefully designed prompt that focuses the models' attention on traffic-relevant elements while minimizing irrelevant background details: *"Please describe this driving scene in detail. Focus on: 1) All visible vehicles (cars, trucks, buses) and their positions; 2) Pedestrians, cyclists, or other traffic participants; 3) Traffic conditions and road layout; 4) Any notable traffic situations or potential hazards. Provide a comprehensive but factual description."*

This prompt explicitly guides models to focus on traffic-related elements rather than background buildings or scenery, requests positional information that can be verified against detection results, and emphasizes factual description to discourage subjective interpretations or speculative content.

*4.1.2.3   Generation Parameter Settings*

We configure the generation parameters to balance diversity and quality. The temperature is set to 0.7 to maintain sufficient randomness for generating diverse captions while avoiding completely random outputs. The maximum token limit is 150 to ensure comprehensive descriptions without excessive verbosity. Each image is processed by each model to generate 10 distinct captions, with different random seeds ensuring variability in the generated content.

## 4.2   VLM-based Hallucination Ground Truth Annotation

While the proposed low-rank methods do not require ground-truth references, we require a validation mechanism to assess the quality of the sorting. To this end, we construct hallucination ground truth labels through an automated pipeline: We construct hallucination detection ground truth by combining NuScenes dataset annotations with large language model-assisted labeling. The truth information sources include official NuScenes 3D object detection bounding boxes with class labels, scene metadata containing weather conditions and time-of-day information, and spatial relationships between detected objects including relative positions and proximity measurements.

Each caption $c_i$ is decomposed into individual sentences using natural language processing techniques:

$$c_i = \{s_{i,j}\}_{j \in \{1,2,\cdots,m_i\}}, i \in \{1,2,\cdots,n\} \tag{9}$$

where $m_i$ is the number of sentences in caption $c_i$.

For each sentence $s_{i,j}$, we conduct a hallucination check using a large VLM: the GPT-4o. The VLM is provided with the original image and some ground truth information, and a structured prompt asking whether the sentence contains any hallucinated content. The ground truth information from NuScenes dataset, for example, includes object detection annotations (3D bounding boxes with precise class labels like vehicles, pedestrians, cyclists, traffic signs and confidence scores) and Scene descriptions (weather conditions, time of day, location context). Based on this information, the VLM evaluates each sentence $s_{i,j}$ based on several criteria including object existence (whether the sentence mentions objects that are not present in the detection results), attribute accuracy (whether the described object attributes such as color, size, and type are consistent with visual evidence) and spatial consistency (whether described spatial relationships match the detected object positions.

Therefore, each sentence $s_{i,j}$ is assigned a binary label indicating whether it contains hallucinations. All results are subsequently verified through manual inspection to ensure accuracy.

We define a binary hallucination indicator:

$$x_{i,j} = \begin{cases} 1, \text{if sentence } s_{i,j} \text{ contrians hallucinations} \\ 0, \text{otherwise} \end{cases} \tag{10}$$

Therefore, for each caption $c_i$, a ground truth hallucination score $h_i^{\text{GT}}$ is evaluated as:

$$h_i^{\text{GT}} = \frac{\sum_{j=1}^{m_i} x_{i,j}}{m_i} \tag{11}$$

This metric captures the proportion of sentences in the caption that are flagged as hallucinated, providing a simple fractional measure where a caption with 4 sentences and 1 hallucinated sentence would receive a score of 0.25. This score serves as the ground truth of hallucination to assess the effectiveness of the proposed low-rank residual analysis.

## 4.3   Evaluation Metrics

Our method addresses a practical problem: identifying and filtering captions with minimal hallucinations from multiple candidates when ground truth is unavailable. We design two evaluation metrics. The first metric emphasizes practical selection performance reflecting application value, and the second focuses on overall sorting capability reflecting theoretical effectiveness.

*4.3.1 Selection Accuracy*

The objective of this evaluation metric is to assess whether selecting the caption with the lowest hallucination score produces the most reliable caption in practice. Since users typically need only the best caption, we evaluate how often our method correctly identifies the truly best caption. For each image, we rank captions by their hallucination scores, select the top-ranked caption, and check if it do not have hallucination. We measure optimal selection accuracy as the percentage of correct selections.

$$A = \frac{1}{m_{i^*}} \sum_{k=1}^{m_{i^*}} x_{i^*,k} \tag{12}$$

where $i^*$ is the index of the caption with lowest hallucination score $i^* = \arg\min_i h_i$.

*4.3.2 Sorting Consistency*

The objective of this evaluation metric is to verify whether our hallucination score $h_i$ sorting aligns with the ground truth hallucination score $h_i^{GT}$ sorting. This is validated by computing Spearman rank correlation coefficients $\rho$ between the estimated hallucination score $h_i$ and ground truth hallucination score $h_i^{GT}$:

$$\rho = \frac{\sum_{i=1}^{n}(R(h_i) - \bar{R}_h)\left(R(h_i^{GT}) - \bar{R}_{h^{GT}}\right)}{\sqrt{\sum_{i=1}^{n}(R(h_i) - \bar{R}_h)^2 \sum_{i=1}^{n}\left(R(h_i^{GT}) - \bar{R}_{h^{GT}}\right)^2}} \tag{13}$$

where $R(h_i)$ and $R(h_i^{GT})$ are the rank of the $i$-th calculated score of $h_i$ and $h_i^{GT}$, $\bar{R}_h$ and $\bar{R}_{h^{GT}}$ are the mean rank of ground truth scores of $h_i$ and $h_i^{GT}$

For each scene, we calculate hallucination scores for all captions using our low-rank residual analysis, simultaneously computing true hallucination scores based on ground truth annotations, and measure sorting consistency using Spearman correlation. We analyze the correlation distribution across all scenes to assess method stability. High sorting consistency indicates our method correctly identifies relative hallucination severity.

## 4.4 Baseline Method: Multi-Agent Debate

To validate the effectiveness of our low-rank approach, we implement a multi-agent debate baseline that represents a popular alternative strategy for hallucination mitigation. This baseline provides a direct comparison point for evaluating both accuracy and computational efficiency.

The multi-agent debate approach operates through iterative refinement across two rounds, where each vision-language model refines its description by considering outputs from other models while re-examining the original image.

**Round 0 (Initial Generation):** Each of the three VLM models generates an initial caption for the driving scene.

**Round 1 (Cross-Model Refinement):** Each model receives its own initial caption alongside the initial captions from the other two models, plus the original scene image. The model is prompted to identify potential hallucinations by checking for objects mentioned in its caption that are absent from other models' descriptions, incorrect spatial relationships or attributes that contradict consensus, and details inconsistent with visual evidence. Each model then generates a refined caption.

**Round 2 (Consensus Alignment):** Each model receives its Round 1 refined caption, the Round 1 refined captions from other models, and again the original scene image. Models are prompted to perform a final verification to ensure no remaining inconsistencies exist, confirm that Round 1 refinements did not introduce new errors, and align descriptions with the emerging multi-model consensus. This produces the final caption for each model.

For model configuration: We maintain identical model configurations and generation parameters as the original caption generation to ensure comparability with our low-rank method. All three models process the same compressed images (400×300 pixels) used throughout our experiments.

## 5 NUSCENES RESULTS

We evaluate the proposed low-rank hallucination mitigation framework under two settings. Section 5.1 is homogeneous multi-agent results: captions are generated by a single VLM using different random seeds. This examines whether the low-rank method is effective for diverse outputs of a single VLM. Section 5.2 is heterogeneous multi-agent results: captions are generated by different VLMs. This investigates whether the low-rank method is effective for outputs of different VLMs.

### 5.1 Homogeneous Multi-agent Results

#### 5.1.1 Selection Accuracy

TABLE 3 summarizes the selection accuracy of each VLM under the homogeneous multi-agent setting, where 10 captions are generated in parallel by the same VLM model, and then apply the low-rank method. The Deepseek model achieves 78% accuracy, indicating that 78% of the selected captions are hallucination-free after applying the low-rank ranking process. Without low-rank analysis, Deepseek's direct outputs contain 71% hallucination-free captions; therefore, our method improves the proportion of accurate descriptions by 7 percentage points. Similar improvements are observed for Llava (from 80% to 81%) and Qwen (from 71% to 73%).

It is important to note that these results are obtained under intentionally challenging conditions—images are compressed, and a high generation temperature is chosen to induce greater hallucinations. This setting helps evaluate the effectiveness of the proposed low-rank method.

TABLE 3 Selection accuracy of 3 VLMs

| Model | Accuracy |
|---|---|
| Single-agent (Deepseek) | 0.71 |
| Single-agent (Llava) | 0.80 |
| Single-agent (Qwen) | 0.71 |
| Homogeneous multi-agent low-rank (Deepseek) | 0.78 |
| Homogeneous multi-agent low-rank (Llava) | 0.81 |
| Homogeneous multi-agent low-rank (Qwen) | 0.73 |

#### 5.1.2 Low-Rank Decomposition Visualization

To provide comprehensive insight into our low-rank decomposition approach, Figure 4 illustrates how the consensus information is separated from hallucinated content. The y-axis represents the amplitude (magnitude) of singular values, which quantifies the importance or "energy" of each component in the matrix decomposition—larger amplitudes indicate components that capture more significant information about the scene. The x-axis shows the singular value index, ordered from largest to smallest.

The plot shows the amplitude distribution of singular values for both the original embedding matrix (black line) and its low-rank approximation (green line). The colored regions provide an intuitive interpretation of our decomposition: the blue "Consensus" region represents the information captured by the low-rank approximation, corresponding to shared patterns across captions that describe genuine scene content. The orange "Hallucination" region highlights the information lost in the low-rank approximation, representing the residual components that correspond to inconsistent or fabricated details. This visualization clearly shows how the low-rank approximation preserves the dominant singular values (consensus information) while discarding smaller singular values that often correspond to caption-specific inconsistencies.

The sharp decay in singular values after the first few components validates our hypothesis that most genuine scene information can be captured in a low-dimensional subspace, while hallucinations manifest as high-dimensional, sparse deviations.

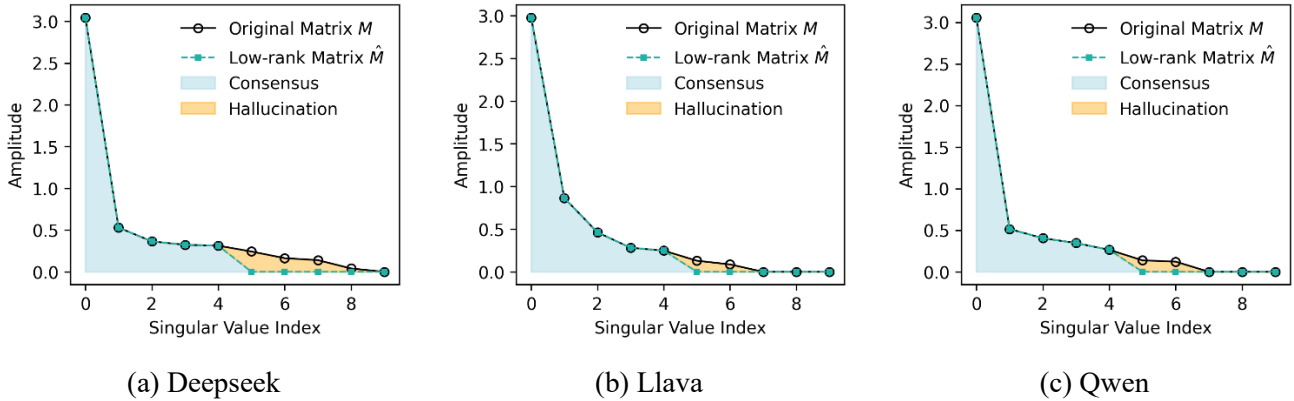

(a) Deepseek  (b) Llava  (c) Qwen

Figure 4 Consensus-hallucination separation (rank=5) using homogeneous multi-agent method

Figure 5 provides a detailed matrix-level view of the decomposition process using heatmap visualizations across the first 40 embedding dimensions for better interpretability (The original embedding matrix **M** has 1024 dimensions). The original matrix **M** displays complex patterns reflecting the diverse semantic content of captions generated by multiple vision-language models. Each row corresponds to one caption, with distinct patterns visible for different models.

The low-rank matrix $\mathbf{R}^*$ preserves the major semantic structures while smoothing out caption-specific variations. The residual matrix $\mathbf{E}^*$ highlights the key insight of our approach: certain rows exhibit significantly higher residual magnitudes, appearing as brighter patterns in the heatmap. These high-residual rows correspond to captions that deviate substantially from the consensus representation, indicating potential hallucinations.

The consistent color scale across all three panels enables direct comparison of magnitude differences. The residual matrix $\mathbf{E}^*$ clearly shows that some captions (particularly visible in rows with higher brightness) contain information patterns that cannot be well-represented by the dominant low-rank subspace, supporting our hypothesis that hallucinated content appears as outliers in the embedding space.

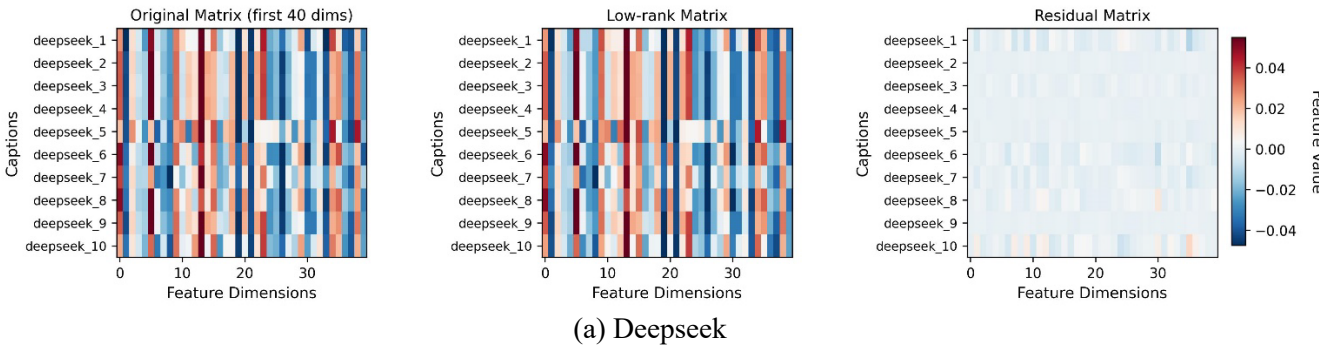

(a) Deepseek

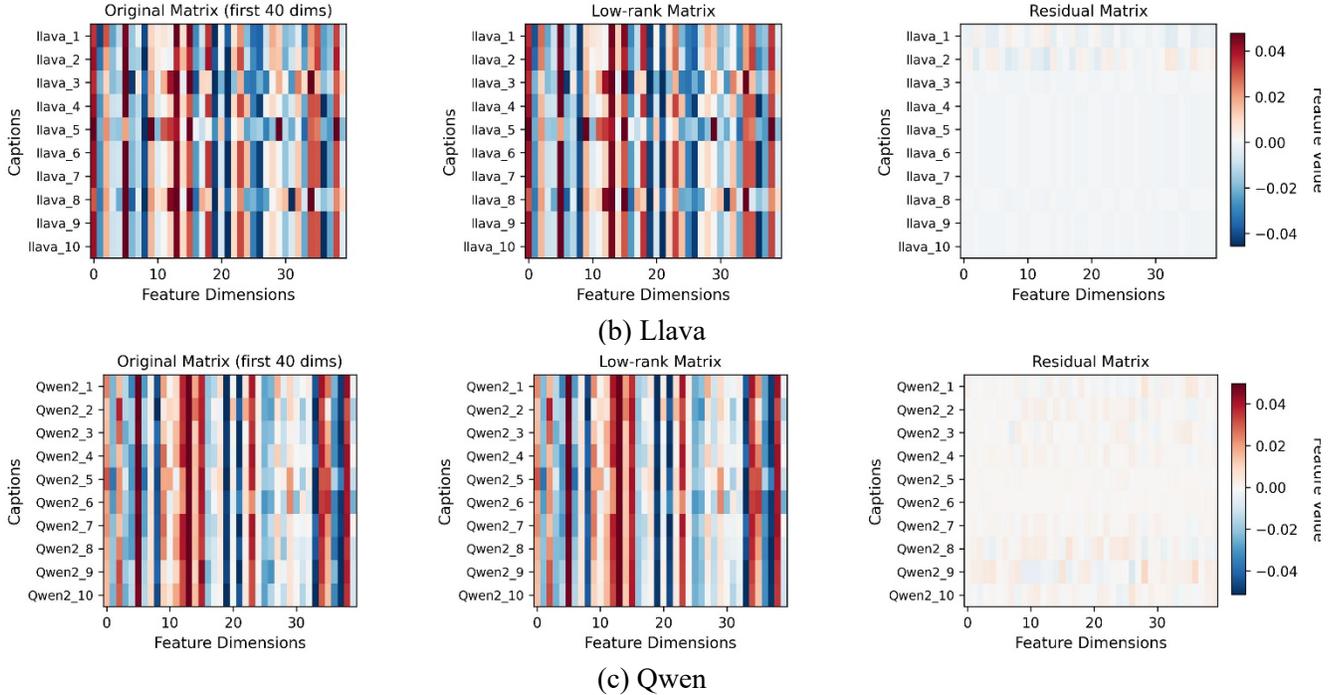

(c) Qwen

Figure 5 Matrix decomposition analysis in homogeneous multi-agent method

### 5.1.3 Sorting Consistency

Our analysis reveals significant variation in sorting consistency across different vision-language models. Figure 6 presents the distribution of Spearman correlation coefficients for all three models across 300 test scenes. The Deepseek model achieves positive correlation in 78% of test cases with an average correlation of 0.52, indicating moderate alignment between our hallucination scores and ground truth sorting. The Qwen model demonstrates improved consistency with positive correlation in 83% of cases and an average correlation of 0.61. Most notably, the Llava model exhibits the strongest performance with positive correlation in 87% of test cases and an average correlation of 0.67.

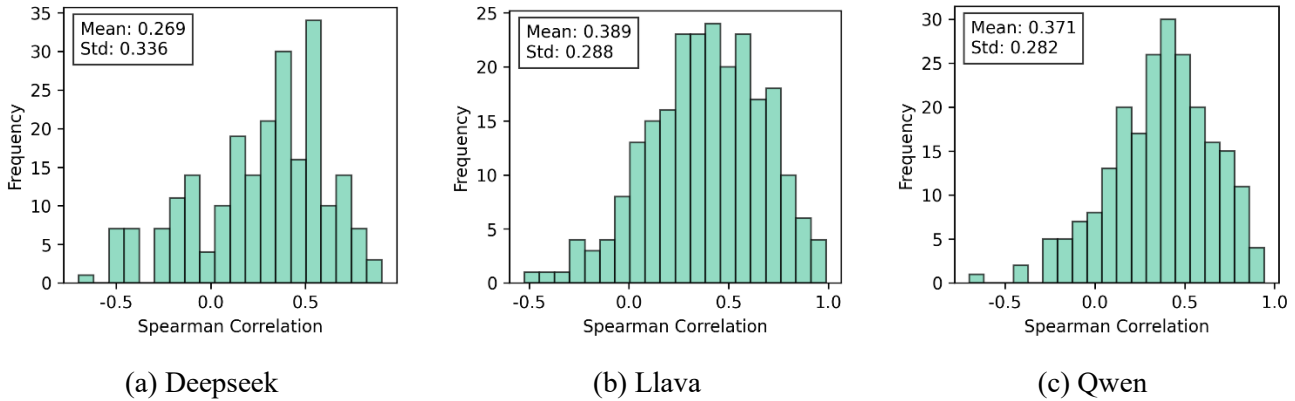

(a) Deepseek  (b) Llava  (c) Qwen

Figure 6 Spearman correlation coefficients for all three models

## 5.2 Heterogeneous Multi-agent Results

In this section, we evaluate the proposed low-rank hallucination mitigation framework under two settings:

1) In the homogeneous setting (Section 5.1): multiple captions are generated by the same VLM using different random seeds or prompts. This configuration examines whether the low-rank method can effectively identify consistent, hallucination-free captions among diverse outputs of a single VLM type; 2) heterogeneous setting (Section 5.2): captions are jointly analyzed from different VLMs (Deepseek, Llava, and Qwen). This configuration further investigates whether integrating diverse VLMs can further mitigate hallucinations.

*5.2.1 Selection Accuracy*

TABLE 4 demonstrates the comparative performance of different hallucination mitigation approaches on top-1 caption selection. Results reveal important insights into the effectiveness of various strategies for identifying the most reliable caption from multiple candidates.

**Heterogeneous Multi-Agent Debate Performance**: The debate baseline shows mixed results across different VLMs' leadership. When Deepseek leads the debate process, accuracy improves to 82%, representing a 4% gain over its single-model performance. Qwen-led debate reaches 81%, showing a substantial 8% improvement from its single-model baseline. These two models successfully refine outputs while maintaining descriptive richness.

However, Llava-led debate achieves only 79% accuracy, a 2% decrease from its single-model performance. This degradation reveals a critical limitation: debate requires models to revise descriptions based solely on other models' text outputs, essentially using uncertain information to correct uncertain information. This often leads to information loss as models adopt overly conservative strategies, preferring to omit details rather than risk disagreement. Additionally, the sequential nature of two-round refinement can compound errors when incorrect descriptions form early consensus.

**Low-Rank Superiority**: Our proposed multi-model low-rank approach achieves 87% accuracy, outperforming all baseline methods. Superior performance stems from fundamental methodological advantages. Unlike the sequential revision process in debate method, our approach doesn't generate new captions but identifies the consensus.

While the improvement over debate appears modest, both methods remain fundamentally limited by the perceptual capabilities of the underlying vision-language models. When all models hallucinate the same object, neither approach can correct the error.

TABLE 4 Selection accuracy comparison across different hallucination mitigation methods

| Method | Accuracy |
|---|---|
| Homogeneous multi-agent low-rank (Deepseek) | 0.78 |
| Homogeneous multi-agent low-rank (Llava) | 0.81 |
| Homogeneous multi-agent low-rank (Qwen) | 0.73 |
| Homogeneous multi-agent debate (lead by Deepseek) | 0.82 |
| Heterogeneous multi-agent debate (lead by Llava) | 0.79 |
| Heterogeneous multi-agent debate (lead by Qwen) | 0.81 |
| Heterogeneous multi-agent low-rank | **0.87** |

*5.2.2 Low-Rank Decomposition Visualization*

Figure 7 shows the singular value decomposition of embedding matrices, where the sharp decay after the first few components confirms that genuine scene information exists in a low-rank matrix $\mathbf{R}^*$ (consensus), and hallucinations appear as high-dimensional sparse deviations in the residual matrix $\mathbf{E}^*$ (hallucination). This validates the proposed approach is valid in heterogeneous multi-agent scenario.

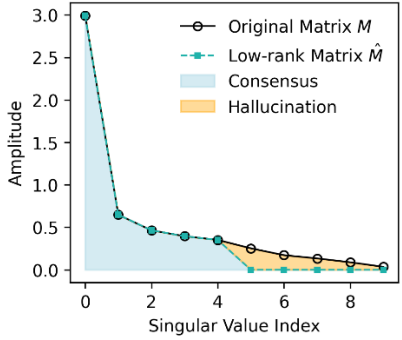

Figure 7 Consensus-Hallucination Separation

Figure 8 provides the matrix-level view of the decomposition process of the first 40 embedding dimensions in multi-agent mode. Each row corresponds to one caption from different VLM models.

The low-rank matrix $\mathbf{R}^*$ is similar to the original matrix but has lower rank, containing the consensus among different VLMs. The residual matrix E highlights that the captions from Qwen model exhibit higher magnitudes, indicating potential hallucinations. These captions contain information patterns that cannot be well-represented by the dominant low-rank subspace, supporting our hypothesis that hallucinated content appears as outliers in the embedding space. Captions from DeepSeek are lighter in color and less hallucinatory.

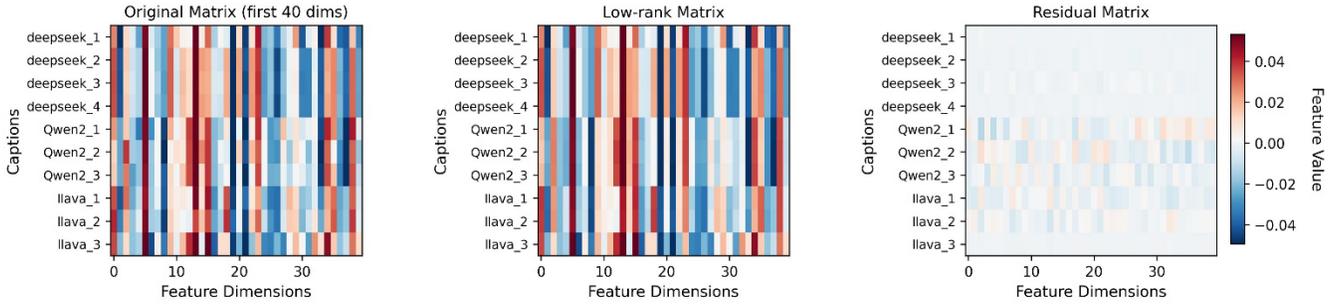

Figure 8 Matrix-Level Decomposition Analysis

### 5.2.3 Sorting Consistency

Combining captions from multiple vision-language models substantially improves sorting consistency. Figure 9 compares multi-model performance against single-model results through correlation distribution analysis. When pooling captions from all three models (10 captions per scene), the proportion of positively correlated scenes increases to 92%, representing improvements of 14%, 9%, and 5% over individual models respectively. The average Spearman correlation coefficient improves to 0.71, exceeding the best single-model performance by 0.04. More importantly, correlation variance decreases by 35%, indicating increased stability across diverse scenarios. This improvement stems from increased caption diversity, which enhances the contrast between consistent and inconsistent descriptions in the embedding space.

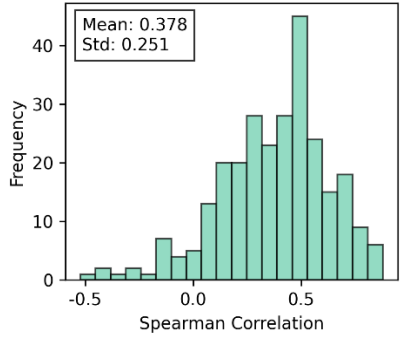

Figure 9 Spearman correlation coefficients for multi-model

To provide a more intuitive understanding of the hallucination detection task, Figure 10 presents a street scene example and the captions generated by different VLMs. Hallucinations in captions including incorrectly described vehicle colors, non-existent objects, or inaccurate spatial relationships. the caption with hallucinations gets higher residual error. Qwen model generates hallucinates of specific vehicle color, resulting in a higher error score (0.1205), while Deepseek model produces more accurate descriptions with lower hallucination rates (0.0496).

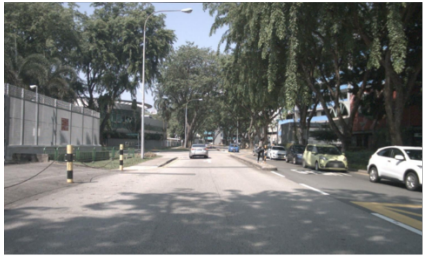

Figure 10 Example of hallucination detection in VLM outputs.

### 5.2.4 Time consuming comparison

To ensure a fair comparison, both the debate and low-rank approaches were executed under identical parallel initialization settings. Specifically, in the debate framework, the three VLM agents, each generated their initial captions in parallel before entering the iterative debate process described in the reference paper. Likewise, in our low-rank approach, the same three agents also generated their initial captions in parallel, after which the low-rank method was applied only once to these outputs.

TABLE 5 compares the computational efficiency of different methods. The multi-model low-rank approach requires only 0.84 seconds per scene, reducing inference time by 51-67% compared to debate methods. This efficiency gap stems from fundamental architectural differences: our approach requires only a single forward pass per model plus efficient SVD computation, while debate methods necessitate multiple inference rounds with full image processing at each iteration. For real-time autonomous driving applications where millisecond-level latency is critical, our method provides both superior accuracy (87% vs 79-82%) and the computational efficiency necessary for practical deployment.

TABLE 5 Inference time comparison across different methods

| Method | Time (s) |
| --- | --- |
| Multi-model debate (lead by Deepseek) | 1.8 |
| Multi-model debate (lead by Llava) | 2.56 |
| Multi-model debate (lead by Qwen) | 1.72 |
| Multi-model low-rank | **0.84** |

## 5.3 Sensitivity analysis

Figure 11 presents a quantitative analysis of how information content is distributed between the low-rank consensus and residual components as a function of the number of retained singular values. The left panel shows the Frobenius norm evolution, where the blue line represents information captured by the low-rank approximation and the red line shows residual information. The intersection point indicates the optimal balance where the low-rank component captures sufficient consensus while the residual effectively isolates outlier patterns.

The right panel displays the cumulative explained variance ratio, demonstrating that approximately 90% of the total variance is captured by the first 5-7 components in typical scenarios. The 80% and 90% threshold lines provide reference points for adaptive rank selection. This analysis validates our adaptive rank selection strategy: by choosing the rank that explains 95% of cumulative variance, we ensure robust consensus extraction while maintaining sensitivity to hallucination detection through the residual component.

The information distribution curves reveal a crucial property of our method: the residual information decreases rapidly initially but then plateaus, indicating that beyond a certain rank threshold, additional components primarily capture noise rather than meaningful semantic variations. This behavior supports the effectiveness of our truncated SVD approach in isolating genuine consensus patterns from hallucinated content.

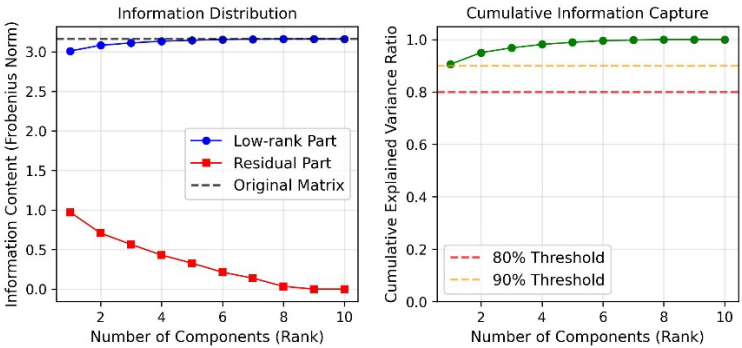

Figure 11 Information Content Distribution Analysis

To understand the semantic structure and variation patterns in multi-agent caption generation, we performed Principal Component Analysis (PCA) on the sentence embeddings of captions generated by three different vision-language models. Figure 12 shows the projection of captions onto the first two principal components, with points colored by their hallucination scores $h_i$. Captions with lower hallucination scores (blue points) tend to cluster together in the principal component space, suggesting semantic consistency among accurate descriptions, and this consistency can be well-captured by the dominant principal components. Captions with higher hallucination scores (red points) appear more dispersed and isolated in space.

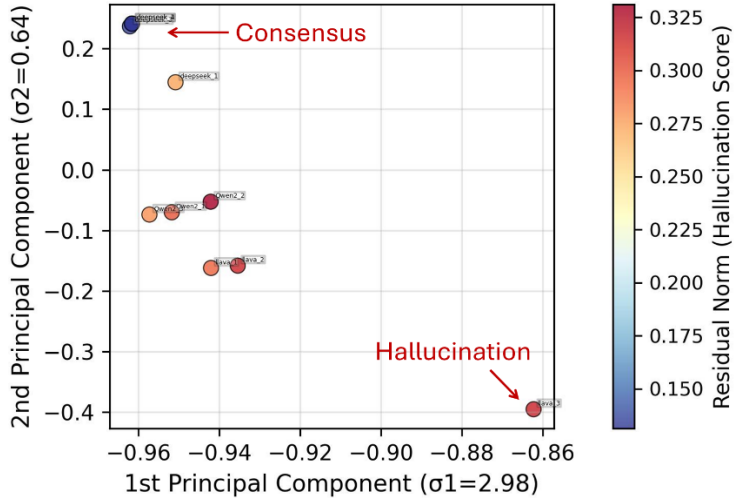

Figure 12 Principal Component Analysis of Multi-Agent Caption Generation

## 6    CONCLUSION AND FUTURE RESEARCH

In this paper, we propose a low-rank method to mitigate hallucinations in VLMs without ground-truth references. By representing multiple candidate captions in an embedding matrix and applying truncated SVD, we decomposed the consensus information from hallucination-prone residuals. Captions with lower residual norms were found to align more closely with ground truth, enabling reliable caption sorting and selection.

Experiments on the NuScenes dataset demonstrated that our approach achieves 87% accuracy in autonomous driving scene captions, representing a 19% improvement over the unfiltered baseline and a 6-10% improvement over the multi-agent debate method. This superior performance in the core task of hallucination-free caption selection is further validated by high sorting consistency with human-verified hallucination labels, as measured by Spearman correlation coefficients. When compared against multi-agent debate baselines, our method not only achieves higher accuracy but also reduces inference time by 51-67%. This combination of accuracy and computational efficiency makes our approach particularly suitable for real-time safety-critical applications such as autonomous driving. These results confirm that low-rank residual analysis provides a simple yet effective strategy for improving the trustworthiness of VLM-generated descriptions in safety-critical applications.

Future research will extend this framework in several directions. First, while we focused on textual hallucination filtering, integrating visual consistency checks may further enhance robustness. Second, applying the method in multi-agent and multi-modal contexts, such as cooperative perception for autonomous driving, could strengthen hallucination detection by exploiting complementary views. Third, incorporating closed-loop simulation with vehicle dynamics models will enable evaluation of how hallucinations propagate into downstream planning and control, bridging the gap between perception reliability and system-level safety.

**ACKNOWLEDGEMENT**

This work was supported by the National Science Foundation Cyber-Physical Systems (CPS) program. Award Number: 2343167.

## REFERENCES

Arai, H., Miwa, K., Sasaki, K., Watanabe, K., Yamaguchi, Y., Aoki, S., Yamamoto, I., 2025. CoVLA:

Comprehensive Vision-Language-Action Dataset for Autonomous Driving, in: Proceedings of the Winter Conference on Applications of Computer Vision (WACV). pp. 1933–1943.

Azaria, A., Mitchell, T., 2023. The Internal State of an LLM Knows When It's Lying. Presented at the The 2023 Conference on Empirical Methods in Natural Language Processing.

Bai, Z., Wang, P., Xiao, T., He, T., Han, Z., Zhang, Z., Shou, M.Z., 2025. Hallucination of Multimodal Large Language Models: A Survey. https://doi.org/10.48550/arXiv.2404.18930

Caesar, H., Bankiti, V., Lang, A.H., Vora, S., Liong, V.E., Xu, Q., Krishnan, A., Pan, Y., Baldan, G., Beijbom, O., 2020. nuScenes: A Multimodal Dataset for Autonomous Driving. Presented at the 2020 IEEE/CVF Conference on Computer Vision and Pattern Recognition (CVPR), pp. 11618–11628. https://doi.org/10.1109/CVPR42600.2020.01164

Chen, C., Liu, K., Chen, Z., Gu, Y., Wu, Y., Tao, M., Fu, Z., Ye, J., 2024. INSIDE: LLMS' INTERNAL STATES RETAIN THE POWER OF HALLUCINATION DETECTION.

Chuang, Y.-S., Xie, Y., Luo, H., Kim, Y., Glass, J., He, P., 2024. DoLa: Decoding by Contrasting Layers Improves Factuality in Large Language Models. https://doi.org/10.48550/arXiv.2309.03883

Cohen, R., Hamri, M., Geva, M., Globerson, A., 2023. LM vs LM: Detecting Factual Errors via Cross Examination. https://doi.org/10.48550/arXiv.2305.13281

Du, X., Xiao, C., Li, Y., 2024. HaloScope: Harnessing Unlabeled LLM Generations for Hallucination Detection. Advances in Neural Information Processing Systems 37, 102948–102972.

Du, Y., Li, S., Torralba, A., Tenenbaum, J.B., Mordatch, I., 2024. Improving factuality and reasoning in language models through multiagent debate, in: Proceedings of the 41st International Conference on Machine Learning, ICML'24. JMLR.org, Vienna, Austria, pp. 11733–11763.

Farquhar, S., Kossen, J., Kuhn, L., Gal, Y., 2024. Detecting hallucinations in large language models using semantic entropy. Nature 630, 625–630. https://doi.org/10.1038/s41586-024-07421-0

Gao, L., Dai, Z., Pasupat, P., Chen, A., Chaganty, A.T., Fan, Y., Zhao, V.Y., Lao, N., Lee, H., Juan, D.-C., Guu, K., 2023. RARR: Researching and Revising What Language Models Say, Using Language Models. https://doi.org/10.48550/arXiv.2210.08726

Guo, Z., Yagudin, Z., Lykov, A., Konenkov, M., Tsetserukou, D., 2024. VLM-Auto: VLM-based Autonomous Driving Assistant with Human-like Behavior and Understanding for Complex Road Scenes, in: 2024 2nd International Conference on Foundation and Large Language Models (FLLM). Presented at the 2024 2nd International Conference on Foundation and Large Language Models (FLLM), pp. 501–507. https://doi.org/10.1109/FLLM63129.2024.10852498

Huang, L., Yu, W., Ma, W., Zhong, W., Feng, Z., Wang, H., Chen, Q., Peng, W., Feng, X., Qin, B., Liu, T., 2025. A Survey on Hallucination in Large Language Models: Principles, Taxonomy, Challenges, and Open Questions. ACM Trans. Inf. Syst. 43, 42:1-42:55. https://doi.org/10.1145/3703155

Jiang, C., Zhang, Y., Cai, Y., Chan, C.-M., Liu, Y., Chen, M., Xue, W., Guo, Y., 2025. Semantic Voting: A Self-Evaluation-Free Approach for Efficient LLM Self-Improvement on Unverifiable Open-ended Tasks. https://doi.org/10.48550/arXiv.2509.23067

Karmanov, A., Guan, D., Lu, S., Saddik, A.E., Xing, E., n.d. Efficient Test-Time Adaptation of Vision-Language Models.

Li, B., Wang, Y., Mao, J., Ivanovic, B., Veer, S., Leung, K., Pavone, M., 2024. Driving Everywhere with Large Language Model Policy Adaptation. Presented at the 2024 IEEE/CVF Conference on Computer Vision and Pattern Recognition (CVPR), IEEE, Seattle, WA, USA, pp. 14948–14957. https://doi.org/10.1109/CVPR52733.2024.01416

Li, K., Patel, O., Viégas, F., Pfister, H., Wattenberg, M., 2023. Inference-Time Intervention: Eliciting Truthful Answers from a Language Model. Advances in Neural Information Processing Systems 36, 41451–41530.

Li, X., Zhao, R., Chia, Y.K., Ding, B., Joty, S., Poria, S., Bing, L., 2023. Chain-of-Knowledge: Grounding Large Language Models via Dynamic Knowledge Adapting over Heterogeneous Sources. Presented at the The Twelfth International Conference on Learning Representations.

Liang, T., He, Z., Jiao, W., Wang, X., Wang, Y., Wang, R., Yang, Y., Shi, S., Tu, Z., 2024. Encouraging Divergent Thinking in Large Language Models through Multi-Agent Debate. https://doi.org/10.48550/arXiv.2305.19118

Lin, Z., Niu, Z., Wang, Z., Xu, Y., 2024. Interpreting and Mitigating Hallucination in MLLMs through Multi-agent Debate. https://doi.org/10.48550/arXiv.2407.20505

Liu, H., Li, C., Li, Y., Lee, Y.J., 2024. Improved Baselines with Visual Instruction Tuning. Presented at the Proceedings of the IEEE/CVF Conference on Computer Vision and Pattern Recognition, pp. 26296–26306.

Long, K., Shi, H., Liu, J., Li, X., 2024. VLM-MPC: Vision Language Foundation Model (VLM)-Guided Model Predictive Controller (MPC) for Autonomous Driving. arXiv preprint arXiv:2408.04821.

Lu, H., Liu, W., Zhang, B., Wang, B., Dong, K., Liu, B., Sun, J., Ren, T., Li, Z., Yang, H., Sun, Y., Deng, C., Xu, H., Xie, Z., Ruan, C., 2024. DeepSeek-VL: Towards Real-World Vision-Language Understanding. https://doi.org/10.48550/arXiv.2403.05525

Peng, B., Galley, M., He, P., Cheng, H., Xie, Y., Hu, Y., Huang, Q., Liden, L., Yu, Z., Chen, W., Gao, J., 2023. Check Your Facts and Try Again: Improving Large Language Models with External Knowledge and Automated Feedback. https://doi.org/10.48550/arXiv.2302.12813

Qian, T., Chen, J., Zhuo, L., Jiao, Y., Jiang, Y.-G., 2024. NuScenes-QA: A Multi-modal Visual Question Answering Benchmark for Autonomous Driving Scenario. https://doi.org/10.48550/arXiv.2305.14836

Wang, P., Bai, S., Tan, S., Wang, S., Fan, Z., Bai, J., Chen, K., Liu, X., Wang, J., Ge, W., Fan, Y., Dang, K., Du, M., Ren, X., Men, R., Liu, D., Zhou, C., Zhou, J., Lin, J., 2024. Qwen2-VL: Enhancing Vision-Language Model's Perception of the World at Any Resolution. https://doi.org/10.48550/arXiv.2409.12191

Wang, X., Wei, J., Schuurmans, D., Le, Q., Chi, E., Narang, S., Chowdhery, A., Zhou, D., 2023. Self-Consistency Improves Chain of Thought Reasoning in Language Models. https://doi.org/10.48550/arXiv.2203.11171

Wang, Z., Mao, S., Wu, W., Ge, T., Wei, F., Ji, H., 2024. Unleashing the Emergent Cognitive Synergy in Large Language Models: A Task-Solving Agent through Multi-Persona Self-Collaboration. https://doi.org/10.48550/arXiv.2307.05300

Yang, Y., Ma, Y., Feng, H., Cheng, Y., Han, Z., 2025. Minimizing Hallucinations and Communication Costs: Adversarial Debate and Voting Mechanisms in LLM-Based Multi-Agents. Applied Sciences 15, 3676. https://doi.org/10.3390/app15073676

Yu, H., Zheng, K., Fang, J., Guo, H., Feng, W., Wang, S., 2018. Co-Saliency Detection Within a Single Image. AAAI 32. https://doi.org/10.1609/aaai.v32i1.12310

Zhang, Yue, Li, Y., Cui, L., Cai, D., Liu, L., Fu, T., Huang, X., Zhao, E., Zhang, Yu, Chen, Y., Wang, L., Luu, A.T., Bi, W., Shi, F., Shi, S., 2025. Siren's Song in the AI Ocean: A Survey on Hallucination in Large Language Models. Computational Linguistics 1–46. https://doi.org/10.1162/COLI.a.16